\documentclass[lettersize,journal]{IEEEtran}
\usepackage{amsmath,amsfonts}
\usepackage{algorithmic}
\usepackage{algorithm}
\usepackage{array}
\usepackage[caption=false,font=normalsize,labelfont=sf,textfont=sf]{subfig}
\usepackage{textcomp}
\usepackage{stfloats}
\usepackage{url}
\usepackage{verbatim}
\usepackage{graphicx}
\usepackage{cite}

\usepackage{soul}
\usepackage{url}
\usepackage[utf8]{inputenc}
\usepackage{graphicx}
\usepackage{amsthm}
\usepackage{booktabs}
\usepackage{algorithm}
\usepackage{algorithmic}
\usepackage[switch]{lineno}
\usepackage{multirow}
\usepackage{verbatim}

\begin{document}

\title{MiM-ISTD: Mamba-in-Mamba for Efficient Infrared Small Target Detection}

\author{Tianxiang Chen, Zi Ye, Zhentao Tan, Tao Gong, Yue Wu, Qi Chu, Bin Liu\\
Nenghai Yu, Jieping Ye, \textit{Fellow, IEEE}
        
\thanks{Tianxiang Chen, Zhentao Tan, Tao Gong, Qi Chu, Bin Liu, Nenghai Yu are with the Key Laboratory of Electromagnetic Space Information, Chinese Academy of Sciences, and the School of Cyber Science and Technology, University of Science and Technology of China, Hefei 230022, China (e-mail: txchen@mail.ustc.edu.cn, tzt@mail.ustc.edu.cn, tgong@.ustc.edu.cn, qchu@ustc.edu.cn, flowice@ustc.edu.cn, ynh@ustc.edu.cn). \\
Zi Ye is with the Institute of Intelligent Software, Guangzhou, China (e-mail: yezi1022@gmail.com).\\
Yue Wu and Jieping Ye are with Alibaba Cloud, Hangzhou, China (e-mail: matthew.wy@alibaba-inc.com, yejieping.ye@alibaba-inc.com).\\ 
Corresponding Author: Tao Gong.}}% <-this % stops a space

\markboth{Journal of \LaTeX\ Class Files,~Vol.~14, No.~8, August~2021}%
{Shell \MakeLowercase{\textit{et al.}}: A Sample Article Using IEEEtran.cls for IEEE Journals}


\maketitle

% 加和visual mmaba，以及有conv在inner mamba block的特征、最终结果可视化对比
% 更详细的说明global context对istd的意义

\begin{abstract}
   Recently, infrared small target detection (ISTD) has made significant progress, thanks to the development of basic models. Specifically, the models combining CNNs with transformers can successfully extract both local and global features. However, the disadvantage of the transformer is also inherited, i.e., the quadratic computational complexity to sequence length. Inspired by the recent basic model with linear complexity for long-distance modeling, Mamba, we explore the potential of this state space model for ISTD task in terms of effectiveness and efficiency in the paper. However, directly applying Mamba achieves suboptimal performances due to the insufficient harnessing of local features, which are imperative for detecting small targets. Instead, we tailor a nested structure, Mamba-in-Mamba (MiM-ISTD), for efficient ISTD. It consists of Outer and Inner Mamba blocks to adeptly capture both global and local features. Specifically, we treat the local patches as "visual sentences" and use the Outer Mamba to explore the global information. We then decompose each visual sentence into sub-patches as "visual words" and use the Inner Mamba to further explore the local information among words in the visual sentence with negligible computational costs. By aggregating the visual word and visual sentence features, our MiM-ISTD can effectively explore both global and local information. Experiments on NUAA-SIRST and IRSTD-1k show the superior accuracy and efficiency of our method. Specifically, MiM-ISTD is $8 \times$ faster than the SOTA method and reduces GPU memory usage by 62.2$\%$ when testing on $2048 \times 2048$ images, overcoming the computation and memory constraints on high-resolution infrared images. 
\end{abstract}


\section{Introduction}
\IEEEPARstart{I}{nfrared} small target detection (ISTD) has been widely applied in remote sensing and military tracking systems. It is a binary segmentation task aiming to segment small target pixels from the background. The task is challenging because the targets are so small that present methods easily miss them or confound them with other background disturbances. %Also, some targets have irregular shapes, which further adds difficulty to the task. %Present ISTD methods still suffer from the above challenges, so our motivation is to tailor a new method for ISTD to address these issues.

Present ISTD methods can be classified into traditional methods and deep-learning-based methods. In the early stages, traditional methods \cite{zhao2021three,zhu2019infrared,bai2018derivative,sun2020infrared,marvasti2018flying,zhang2019infrared,han2019local,rivest1996detection,chen2013local} take the dominance. However, these methods rely on prior knowledge and handcraft features, resulting in limited accuracy when applied to images that do not conform to their assumptions. 

\begin{figure*}
    \centering    \includegraphics[width=0.99\textwidth,height=0.29\textwidth]{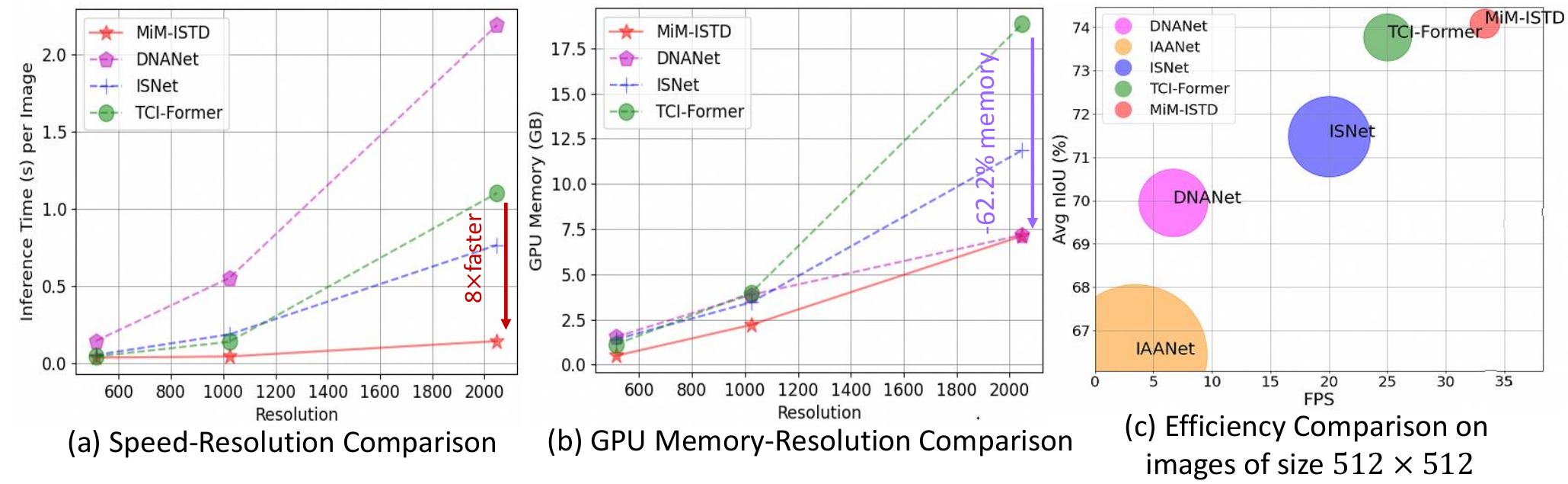}
    \caption{(a), (b) MiM-ISTD is more computation and memory efficient than present SOTA methods, DNANet \cite{li2022dense} and TCI-Former \cite{Chen_Tan_Chu_Wu_Liu_Yu_2024}, in dealing with high-resolution infrared images. Specifically, MiM-ISTD is $10 \times$ faster than TCI-Former \cite{Chen_Tan_Chu_Wu_Liu_Yu_2024} and saves 62.2$\%$ GPU memory per image with a resolution of $2048 \times 2048$. (c) The overall efficiency comparison on images of resolution $512 \times 512$, where larger bubbles denote higher GPU memory usage.}\label{efficiency}
\end{figure*}

%(c) MiM-ISTD is more robust to resolution increase and still maintains a higher accuracy by comparing the test accuracy with TCI-Former, both trained and tested on $1024 \times 1024$ images resized from original dataset images.

In recent years, deep-learning-based methods have significantly improved ISTD performances, and most of them are CNN-based methods \cite{zhao2020novel,chen2023bauenet,dai2021attentional,zhang2023attention,dai2021asymmetric,wang2019miss,li2022dense,zhang2022isnet}. However, the drawback of CNN-based methods is that their focus on local features is at the cost of global contexts. Global contexts are also important to ISTD because in infrared images the background pixels and the small targets seem so similar in many cases that cannot be distinguished by local features alone, which easily triggers missed detection. As a solution, some hybrid methods \cite{Chen_Tan_Chu_Wu_Liu_Yu_2024,chen2023abmnet,wang2022interior,qi2022ftc,chen2023fluid,liu2021infrared,zhang2022rkformer} that combine ViT with CNN have been proposed to rely on ViT's ability to model long-range dependencies. However, these methods generally suffer from a heavier computational burden due to the quadratic computational complexity of ViT. Despite certain work \cite{chen2023abmnet} adopting linear ViTs, its accuracy is still subordinate to the designs with quadratic complexity. Considering that high-resolution images are not rare in the infrared remote sensing domain (e.g. images produced by high-resolution infrared military sensors), this efficiency defect will be amplified when the resolution gets larger and hinder real-time ISTD. How to relieve the inefficiency while maintaining high ISTD accuracy is our main concern.

Recently, State Space Models (SSMs) have drawn increasing interest among researchers. Mamba \cite{gu2023mamba} is the first proposed basic model built by SSMs and has achieved promising performance compared to booming transformers in various long-sequence modeling tasks while maintaining a linear complexity. In a short time, Mamba has achieved success in various fields \cite{gu2021efficiently,ma2024u,liu2024vmamba,ye2024p} and is considered to have the potential to become the next-generation basic model after transformers. However, when we directly transfer visual Mamba \cite{liu2024vmamba} to ISTD, the detection accuracy is not high, despite impressive model efficiency. The reason is that in ISTD the targets are typically very small, necessitating a greater emphasis on local features compared to other vision tasks that predominantly involve standard-size targets. Unfortunately, Mamba is not proficient in capturing these critical local features.

We aim to propose a Mamba-based ISTD encoder to solve the locality defect while still maintaining a superior model efficiency. To this end, we get inspired by TNT \cite{han2021transformer}, which effectively models local structural features with a trivial increase of computation and memory cost, and propose a novel Mamba-in-Mamba (MiM-ISTD) architecture for more efficient ISTD. To boost the feature representation ability of Mamba, we divide the input image into several patches as "visual sentences" and then further divide them into sub-patches as "visual words". We use an Outer Mamba block to extract features of visual sentences, and further assist it with Inner Mamba blocks to excavate the local features of smaller visual words. In particular, features and relations between visual words in each visual sentence are calculated independently using a shared network to ensure the added amount of parameters and FLOPs is minimal. Then, the visual word features are consolidated back into their respective sentences. In this way, MiM-ISTD enables us to extract visual information with refined granularity. We present in Fig.~\ref{efficiency} that MiM-ISTD exhibits superior efficiency over other methods in terms of GPU memory usage and inference time, particularly as the resolution of infrared images increases. In general, MiM-ISTD can achieve the most notable accuracy-efficiency balance compared with other SOTA methods. %We then conduct a series of experiments on two widely used ISTD datasets, NUAA-SIRST and IRSTD-1k, to demonstrate its superiority and thoroughly analyze the impact of the size for dividing visual words. The results show that our TNT can achieve better accuracy and efficiency trade-off over other state-of-the-art transformer methods.

Our contributions can be summarized in three folds:
\begin{itemize}
    \item To the best of our knowledge, we are the first to apply Mamba to ISTD successfully, providing a new benchmark and valuable insights for future advancements in efficient and potent Mamba-based methods.
    \item To apply Mamba to the ISTD domain, we tailor a Mamba-in-Mamba (MiM-ISTD) structure in order to guarantee higher efficiency while sufficiently extracting both local and global information.
    \item Experiments on two public ISTD datasets, NUAA-SIRST and IRSTD-1k, prove the superior accuracy and efficiency of our method. Specifically, MiM-ISTD achieves a speedup of 8 times over the SOTA method while also cutting down GPU memory usage by 62.2$\%$ for each $2048 \times 2048$ image during inference.
\end{itemize}

\section{Related work}
\subsection{Infrared Small Target Detection Networks}
Generally speaking, present ISTD networks can be classified into two categories: CNN-based and hybrid networks. CNN-based networks mainly focus on local feature extraction. Dai et al. \cite{dai2021asymmetric} propose asymmetric contextual modulation (ACM) for cross-layer information exchange to improve ISTD performance. 
They also design AlcNet \cite{dai2021attentional}, including a local attention module and a cross-layer fusion module to preserve the local features of small targets. Wang et al. propose MDvsFA \cite{wang2019miss}, which applies generative adversarial network (GAN) to ISTD, and achieves a trade-off between miss detection and false alarm. BAUENet \cite{chen2023bauenet} introduces uncertainty to ISTD and achieves boundary-aware segmentation. DNANet \cite{li2022dense} progressively interacts with high and low-level features. Dim2Clear \cite{zhang2023dim2clear} treats ISTD as an image detail reconstruction task by exploring the image enhancement idea. FC3-Net \cite{zhang2022exploring} explores feature compensation and cross-level correlation for ISTD task. ISNet \cite{zhang2022isnet} designs a simple Taylor finite difference-inspired block and a two-orientation attention aggregation module to detect targets. Recently, the first diffusion model for ISTD, DCFR-Net\cite{fan2024diffusion}, has been proposed. However, its accuracy is subordinate to the present SOTA method \cite{Chen_Tan_Chu_Wu_Liu_Yu_2024}, and its computational efficiency lags considerably when compared to the majority of deep learning-based ISTD techniques.

Relying solely on local features for ISTD may lead to missed detection of small targets, which can merge into similar backgrounds and become indistinguishable. Therefore, hybrid methods complement local details with global contexts by combining ViT with CNN. IRSTFormer\cite{chen2022irstformer} adopts hierarchical ViT to model long-range dependencies but lays insufficient emphasis on mining local details.
ABMNet \cite{chen2023abmnet} adopts ODE methods in both CNN and linear ViT structure design for ISTD. IAANet \cite{wang2022interior} concatenates local patch outputs from a simple CNN with the original transformer, but causes limited feature extraction, especially in low-contrast scenarios. RKformer \cite{zhang2022rkformer} applies the Runge-Kutta method to build coupled CNN-Transformer blocks to highlight infrared small targets and suppress background interference. TCI-Former \cite{Chen_Tan_Chu_Wu_Liu_Yu_2024} extracts small target features by simulating the thermal conduction process and achieves SOTA results. However, most of these hybrid methods suffer from a quadratic computational complexity due to the usage of ViT. Despite some work \cite{chen2023abmnet} adopting linear ViT design, its detection accuracy cannot be on par with other works with quadratic-complexity ViTs.

To improve network efficiency while maintaining high accuracy, we draw inspiration from both Mamba \cite{gu2023mamba} and TNT \cite{han2021transformer}, and propose a Mamba-in-Mamba architecture for ISTD, within which contains a Mamba-in-Mamba (MiM) hierarchical encoder that implements efficient local and global feature extraction.

\subsection{Mamba in Vision Tasks}
Recently, state space sequence models (SSMs) \cite{gu2021efficiently} have shown promise in efficiently handling long sequence modeling, offering an alternative for addressing long-range dependencies in visual tasks. Compared with transformers, SSMs are more efficient since they scale linearly with sequence length, and retain a superior ability to model long-range dependencies. Several latest studies have demonstrated the effectiveness of Mamba in vision tasks \cite{zhang2024survey}. For instance, Vim \cite{zhu2024vision} proposed a generic vision backbones with bidirectional Mamba blocks. VMamba \cite{liu2024vmamba} proposed a hierarchical Mamba-based vision backbone and a cross-scan module to address the issue of direction-sensitivity arising from the disparities between 1D sequences and 2D image representations.

Notably, Mamba has been most widely applied to the medical image segmentation areas. U-Mamba \cite{ma2024u}, Vm-unet \cite{ruan2024vm}, Mamba-unet \cite{wang2024mamba} and SegMamba \cite{xing2024segmamba} proposed a task-specific architecture with the Mamba block based on nnUNet \cite{isensee2021nnu}, Swin-UNet \cite{cao2022swin}, VMamba \cite{liu2024vmamba} and Swin-UNETR \cite{hatamizadeh2021swin}, respectively. P-Mamba \cite{ye2024p} combined PM diffusion with Mamba to efficiently remove background noise while preserving target edge details. Swin-UMamba \cite{liu2024swin} verified that ImageNet-based pre-training is important to medical image segmentation for Mamba-based networks. %Weak-Mamba-UNet \cite{wang2024weak} and Semi-Mamba-UNet \cite{wang2024semi} are for weak-supervised and semi-supervised medical image segmentation. 
Vivim \cite{yang2024vivim} introduced Mamba for medical video object segmentation.

\begin{figure*}
    \centering    \includegraphics[width=0.99\textwidth,height=0.22\textwidth]{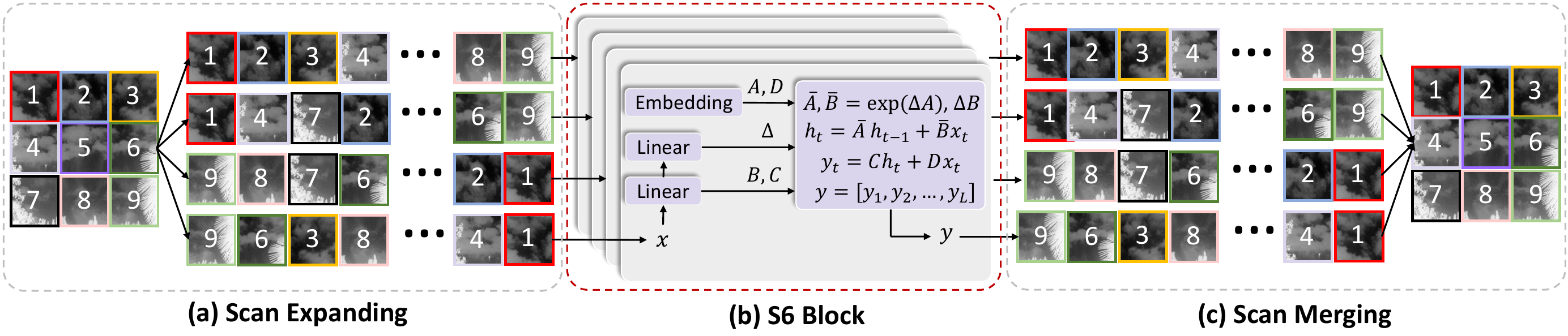}
    \caption{Illustration of the 2D Selective Scan (SS2D) on an infrared image. We commence by scanning an image
using scan expanding. The four resulting feature sequences are then individually processed through the S6
block and the four output sequences are merged (scan merging) to construct the final 2D feature map.}\label{ss2d}
\end{figure*}

Since these Mamba models have achieved promising results in various vision tasks, we intend to study whether Mamba can also 
bring advancements to ISTD as it has brought to other vision tasks. However, when we directly apply visual Mamba blocks to ISTD, the detection accuracy is not considerable, despite superior model efficiency. The reason is that the targets in ISTD tasks are very small, which requires paying more attention to the local features than other vision tasks with mainly common-size targets, while the original visual Mamba block cannot well explore these local features. To solve this defect, we propose Mamba-in-Mamba (MiM-ISTD) for ISTD, which takes visual sentences and visual words flows simultaneously and sends each flow to respective visual Mamba blocks to obtain both local and global features with high efficiency.

\section{Proposed method}
%We illustrate the structure of our MiM-ISTD in \cref{overview}. It mainly consists of 1) a convolutional stem to produce patches and sub-patches, 2) a MiM hierarchical encoder to extract multi-scale global and local features, and 3) a plain decoder to get prediction masks. More details will be described in the following. 
In this section, we describe our proposed Mamba-in-Mamba for efficient ISTD and analyze the computation complexity in detail.

\subsection{Preliminaries}
\subsubsection{State Space Models.} State Space Models (SSMs) are commonly employed as linear time-invariant systems that transform a one-dimensional input stimulus $x(t) \in \mathbb{R}^{L}$ through intermediary implicit states $h(t) \in \mathbb{R}^{N}$ to an output $y(t) \in \mathbb{R}^{L}$. In mathematical terms, SSMs are typically described by linear ordinary differential equations (ODEs) (\eqref{eq:ssm}), where the system is characterized by a set of parameters including the state transition matrix $A \in \mathbb{C}^{N \times N}$, the projection parameters $B,C \in \mathbb{C}^{N}$, and the skip connection $D \in \mathbb{C}^{1}$.

\begin{equation}\label{eq:ssm}
\begin{split}
&h^{'}(t)=Ah(t)+Bx(t)\\
&y(t)=Ch(t)+Dx(t)\\
\end{split}
\end{equation}

% \begin{equation}\label{eq:ssm}
% h^{'}(t)=Ah(t)+Bx(t), \quad y(t)=Ch(t)+Dx(t)\\
% \end{equation}

\subsubsection{Discretization.} State Space Models (SSMs) present significant challenges when applied to deep learning scenarios due to their continuous-time nature. To address this, the ODE needs to be transformed into a discrete function. %The transformation is essential for synchronizing the model with the sample rate of the underlying signal embodied in the input data, facilitating computationally efficient operations [16]. 
Considering the input $x_{k} \in \mathbb{R}^{L \times D}$, a sampled vector within the signal flow of length $L$ following \cite{gupta2022diagonal}, a timescale parameter $\Delta$ can be introduced to the continuous parameters $A$ and $B$ into their discrete counterparts $\overline{A}$ and $\overline{B}$ following the zeroth-order hold (ZOH) rule. Consequently, \eqref{eq:ssm} can be discretized as follows:

\begin{equation}\label{eq:discrete}
\begin{split}
&h_{k}=\overline{A}h_{k-1}+\overline{B}x_{k}, \quad y(t)=Ch(t)+Dx(t)\\
&\overline{A}=e^{\Delta A}, \quad \overline{B}=(e^{\Delta A}-I)A^{-1}B, \quad \overline{C}=C\\
\end{split}
\end{equation}
where $B,C \in \mathbb{R}^{D \times N}$ and $\Delta \in \mathbb{R}^{D}$. In practice, we refine the approximation of $\overline{B}$ using the first-order Taylor series:

\begin{equation}\label{eq:taylor}
\overline{B}=(e^{\Delta A}-I)A^{-1}B=(\Delta A)(\Delta A)^{-1}\Delta B=\Delta B
\end{equation}

\subsubsection{2D Selective Scan.} %The distinction between 2D visual data and 1D language sequences makes it unsuitable to directly apply Mamba to vision tasks. For instance, while 2D spatial information is vital in vision tasks, it is not the primary focus in 1D sequence modeling, resulting in restricted receptive fields that overlook potential relationships with patches unscanned. The 2D selective scan (SS2D) in \cite{liu2024vmamba} can address the issue. 

2D spatial information cannot be effectively captured by models designed for 1D data, making it unsuitable to directly apply Mamba to vision tasks. The 2D selective scan (SS2D) in \cite{liu2024vmamba} can address the issue. The overview of SS2D is depicted in \eqref{eq:ss2d}. SS2D arranges image patches in four different directions to generate four separate sequences. The quad-directional scanning strategy ensures that each element in the feature map integrates information from all other locations in different directions, thus creating a global receptive field without increasing linear computational complexity. Each resulting feature sequence is then processed using the selective scan space state sequential model (S6) before merging to reconstruct the 2D feature map. Given the input feature $z$, then the output feature $\overline{z}$ of SS2D can be expressed as:

\begin{equation}\label{eq:ss2d}
\begin{split}
&z_{i}=expand(z,i)\\
&\overline{z}_{i}=S6(z_{i})\\
&\overline{z}=merge(\overline{z}_{1},\overline{z}_{2},\overline{z}_{3},\overline{z}_{4})\\
\end{split}
\end{equation}
% \begin{equation}\label{eq:ss2d}
% z_{i}=expand(z,i), \quad \overline{z}_{i}=S6(z_{i}), \quad \overline{z}=merge(\overline{z}_{1},\overline{z}_{2},\overline{z}_{3},\overline{z}_{4})\\
% \end{equation}
where $i \in \{1, 2, 3, 4\}$ represents one of the four scanning directions. $expand(·)$ and $merge(·)$ refer to the scan expanding and scan merging operations in \cite{liu2024vmamba}. The S6 block in \eqref{eq:ss2d} enables each element in a 1D array to engage with any previously scanned samples through a condensed hidden state. For a more comprehensive understanding of S6, \cite{liu2024vmamba} provides an in-depth explanation.

\begin{figure*}
    \centering  \includegraphics[width=0.98\textwidth,height=0.53\textwidth]{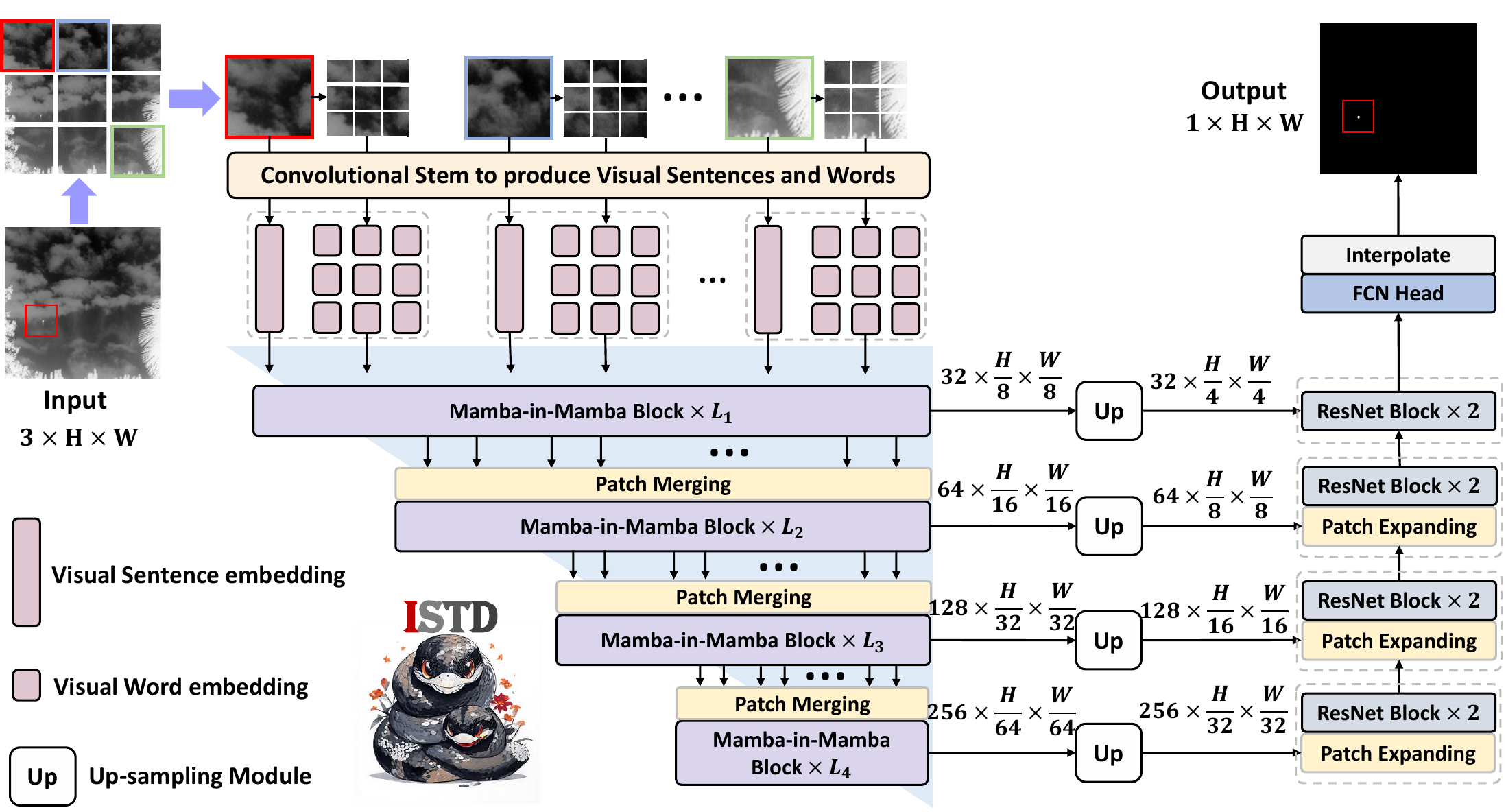}
    \caption{Overview of our MiM-ISTD, which mainly includes a convolutional stem, a pure Mamba-based MiM hierarchical encoder, and a plain decoder.
}\label{overview}
\end{figure*}

\subsection{Mamba-in-Mamba for Efficient ISTD}
Present ISTD methods mainly typically employ CNNs or hybrids of CNNs and ViTs. The latter compensates for the shortcomings of the former in modeling long-range dependencies but suffers from a quadratic computational complexity. Recently, Mamba has been proposed. It is renowned for its superior model efficiency, less GPU memory usage, and better long-range dependency modeling and has been successfully applied to the vision domain. Therefore, we explore whether Mamba can also be applied to improve ISTD performances. However, when we directly apply visual mamba block to ISTD, the accuracy is not very impressive because local features, which matter a lot to detect small targets, are less explored. To address this deficiency, we propose a Mamba-in-Mamba (MiM-ISTD) architecture, shown in Fig.~\ref{overview}, to learn both global and local information in an image while guaranteeing superb model efficiency.

Given a 2D infrared image $\mathcal{X} \in \mathbb{R}^{H \times W \times 3}$, it is divided evenly into $n$ patches to form $\mathcal{X} = [X^1, X^2, ..., X^n]$ where each patch is in $\mathbb{R}^{n \times p \times p \times 3}$, with $(p, p)$ denoting the resolution of each patch. In MiM-ISTD, we view the patches as visual sentences that represent the image. Further, each patch is segmented into $m$ smaller sub-patches, making each visual sentence a sequence of visual words:
\begin{equation}\label{eq:x}
X^i=[x^{i,1},x^{i,2},...,x^{i,m}],j=1,2,...,m
\end{equation}
where $x^{i,j} \in \mathbb{R}^{s \times s \times 3}$ is the $j-th$ visual word of the $i-th$ visual sentence $X^i$, $(s, s)$ is the spatial size of sub-patches. Since our MiM adopts a hierarchical encoder structure, the spatial shapes of visual sentences and words are unfixed and will gradually decrease as the network layers deepen.

\subsubsection{Convolutional Stem.}
We construct a convolutional stem, where a stack
of $3 \times 3$ convolutions is utilized, to produce visual words $\in \mathbb{R}^{\frac{H}{2} \times \frac{W}{2} \times C}$ and visual sentences $\in \mathbb{R}^{\frac{H}{8} \times \frac{W}{8} \times D}$ at the first stage, where $C$ is the visual word dimension and $D$ is the visual sentence dimension. Each visual word corresponds to a $2\times2$ pixel region in the original image, and each visual sentence is composed of $4\times4$ visual words. Unlike ViTs, we do not add the position embedding bias to visual words and sentences due to the causal nature of visual mamba block \cite{liu2024vmamba}. %The word-level and sentence-level position encodings are added on visual words and sentences respectively to preserve spatial information.

\begin{figure*}
    \centering    \includegraphics[width=0.99\textwidth,height=0.20\textwidth]{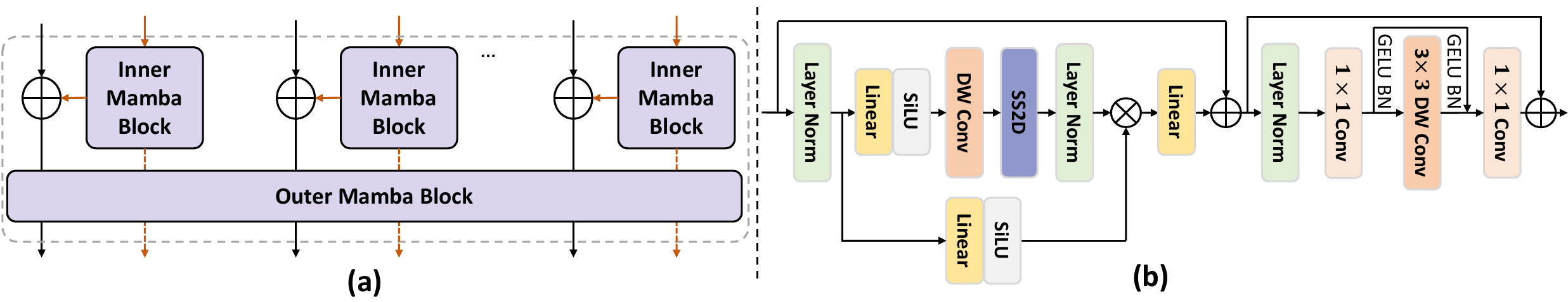}
    \caption{Overview of (a) our proposed Mamba-in-Mamba (MiM) block, which contains an Inner Mamba block and an Outer Mamba block, and (b) the structure of Inner/Outer Mamba block \cite{liu2024vmamba}. The Inner Mamba block is shared in the same layer. The dashed line in (a) means bypassing the Outer Mamba block.}\label{mim_block}
\end{figure*}

\subsubsection{MiM Hierarchical Encoder.} The core part of our MiM is its hierarchical encoder of four stages with different numbers of tokens, as shown in Fig.~\ref{overview}. Across the four
stages, the spatial shape of visual words is set as $\frac{H}{2} \times \frac{W}{2}$, $\frac{H}{4} \times \frac{W}{4}$, $\frac{H}{8} \times \frac{W}{8}$ and $\frac{H}{16} \times \frac{W}{16}$. The spatial shape of visual sentences are set as $\frac{H}{8} \times \frac{W}{8}$, $\frac{H}{16} \times \frac{W}{16}$, $\frac{H}{32} \times \frac{W}{32}$ and $\frac{H}{64} \times \frac{W}{64}$. We adopt the patch merging in \cite{cao2022swin} as the down-sampling operation. Each stage consists of multiple MiM blocks which process both word-level and sentence-level features. The visual words $x^{i,j}$ are mapped to a sequence of word embeddings $w^{i,j}$ via a linear projection, expressed as:
\begin{equation}\label{eq:y}
W^i=[w^{i,1},w^{i,2},...,w^{i,m}],w^{i,j}=FC(Vec(x^{i,j}))
\end{equation}
where $w^{i,j}\in \mathbb{R}^{c}$ is the $j-th$ word embedding of the $i-th$ visual sentence, $c$ is the dimension of word embedding, $W^i$ is the collection of word embeddings of the $i-th$ visual sentence, and $Vec(·)$ refers to the vectorization operation.

The MiM block handles two different data streams: one traverses through the visual sentences, while the other manages the visual words within each sentence. For the word embeddings, the relation between visual words should be exploited as follows:
\begin{equation}\label{eq:inner}
\begin{split}
%W^i_l=W^i_{l-1}+Mamba(LN(W^i_{l-1})), l = 1,2,...,L
&W^{i'}_l=W^i_{l-1}+Mamba(LN(W^i_{l-1}))\\
&W^{i}_l=W^{i'}_l+Conv_{FFN}(LN(W^{i'}_l))\\
\end{split}
\end{equation}
where $l$ represents the index corresponding to the $l^{th}$ block within a series spanning from $1$ to $L$, with $L$ denoting the aggregate number of such blocks. The first word embedding input $W^i_0$ is the $W^i$ in \eqref{eq:y}. $W^{i'}_l$ denotes the intermediate feature. All transformed word embeddings are denoted by $\mathcal{W}_l=[W^{1}_l,W^{2}_l,...,W^{n}_l]$. This can be viewed as an Inner Mamba block, illustrated in Fig.~\ref{mim_block} (a). In this process, the relationships among visual words are built by computing interactions among each two visual words. For example, in a patch containing a small target, a word denoting the target would have a stronger relation with other target-related words while interacting less with the background part.

At the sentence level, we generate sentence embedding memories as storage for the sequence of sentence-level representations, initialized as zero: $\mathcal{S}_0=[S^1_0,S^2_0,...,S^n_0]\in \mathbb{R}^{n \times d}$. In each layer, the sequence of word embeddings is mapped to the domain of sentence embedding by linear projection and subsequently integrated into the sentence embedding:
\begin{equation}\label{eq:convert}
S^i_{l-1}=S^i_{l-1}+FC(Vec(W^i_{l})), l = 1,2,...,L
\end{equation}
By doing so, the sentence embedding can be augmented by the word-level features. We use another Mamba block, denoted as Outer Mamba block, to transform the sentence embeddings:
%\begin{equation}\label{eq:outer}
%\mathcal{S}_{l}=\mathcal{S}_{l-1}+Mamba(LN(\mathcal{S}_{l-1})), l = 1,2,...,L
%\end{equation}

\begin{equation}\label{eq:outer}
\begin{split}
%W^i_l=W^i_{l-1}+Mamba(LN(W^i_{l-1})), l = 1,2,...,L
&\mathcal{S}^{'}_{l}=\mathcal{S}_{l-1}+Mamba(LN(\mathcal{S}_{l-1}))\\
&\mathcal{S}_{l}=\mathcal{S}^{'}_{l}+Conv_{FFN}(LN(\mathcal{S}^{'}_{l}))\\
\end{split}
\end{equation}
The Outer Mamba block can model the relationships among sentence embeddings. The inputs and outputs of the MiM block are visual word embeddings and sentence embeddings, so the MiM hierarchical encoder can be defined as
\begin{equation}\label{eq:mim}
\mathcal{W}_{l},\mathcal{S}_{l}=MiM(\mathcal{W}_{l-1},\mathcal{S}_{l-1}), l = 1,2,...,L
\end{equation}
Within our MiM block, the visual mamba block \cite{liu2024vmamba} is employed in both the Inner and Outer Mamba blocks, each followed by a coupling with a convolutional feed-forward network, as shown in Fig.~\ref{mim_block} (b). The Inner Mamba block models the relationship between visual words for local feature extraction, while the Outer Mamba block captures the global feature by modelling the relationship between visual sentences. The convolutional feed-forward network is responsible for augmenting these features with finer local details. We set the block number $L_1,L_2,L_3,L_4=2,2,2,2$ of each stage by default. Stacking the MiM blocks for $L=8$ layers, we build our MiM hierarchical encoder. 

We feed the visual sentence embedding outputs of each stage to the respective decoder stages. The spatial shapes of the four-stage encoder outputs are $\frac{H}{8} \times \frac{W}{8}$, $\frac{H}{16} \times \frac{W}{16}$, $\frac{H}{32} \times \frac{W}{32}$ and $\frac{H}{64} \times \frac{W}{64}$, which are different from the feature map scales from typical backbones. To solve this discrepancy, these outputs should pass through straightforward up-sampling modules before reaching the decoder. The module includes a $2 \times 2$ transposed convolution with a stride of two, succeeded by batch normalization \cite{ioffe2015batch}, GeLU \cite{hendrycks2016gaussian}, a stride-one $3 \times 3$ convolution, another batch normalization and GeLU. Therefore our MiM hierarchical encoder can produce feature maps with strides of 4, 8, 16, and 32 pixels relative to the input image.

\subsubsection{Decoder Structure.}
In contrast to the patch merging \cite{cao2022swin} operation used in the encoder for down-sampling, we use the patch expanding layer \cite{cao2022swin} in the decoder for up-sampling. In each decoder stage, we integrate the up-sampled encoder outputs with the expanded decoder features and send the integrated feature to basic ResNet blocks. Finally, the features go through a fully connection network head and an interpolation operation to get the final mask prediction.

\subsubsection{Complexity Analysis.}
%Given a visual sequence $T \in \mathbb{R}^{1 \times n \times d}$, the computation complexity of SSM is $8ndN$ \cite{zhu2024vision}, where $N$ is the SSM dimension and is set to 16 by default. For simplicity, we only consider the core SSM in SS2D and 3 linear layers and calculate the FlOPs of a visual Mamba block as $8ndN+3nd^2$. The MiM block includes an inner Mamba block, an outer Mamba block, and a linear layer, with complexities of $8mncN+3mnc^2$, $8ndN+3nd^2$, and $nmcd$, where $m$ is the number of visual words in a visual sentence and $c$ is the word embedding dimension. Thus, the total FLOPs for the MiM block $FLOPs_{MiM}$ sum to $8mncN+8ndN+3mnc^2+3nd^2+nmcd$, still maintaining a linear complexity. While the FLOPs of the standard transformer block $FLOPs_{ST}$ is $2nd(6d+n)$ \cite{han2021transformer}. Considering that $c \ll d$ and $m \ll n$ in high resolution infrared images, the ratio of $FLOPs_{MiM}$ and $FLOPs_{ST}$ approaches 0, meaning that our MiM block introduces a trivial FLOP increase while offering a superior accuracy-efficiency balance demonstrated in subsequent experiments.

Given a visual sequence $T \in \mathbb{R}^{1 \times n \times d}$, the computation complexity of SSM is $3nEN + nEN = 96nd + 32nd = 128nd$ \cite{zhu2024vision}, where $N$ is the SSM dimension and is set to 16 by default and $E=2d$. In comparison, the computation complexity of self-attention is $4nd^2 + 2n^{2}d$. It is evident that self-attention exhibits quadratic complexity to the sequence length $n$, whereas SSM is linear. This computational efficiency makes our MiM-ISTD more scalable compared to other quadratic Transformer-based models like IAANet \cite{wang2022interior} and RKformer \cite{zhang2022rkformer}. 

Each of our proposed MiM block includes 16 Inner Mamba block and an Outer Mamba block, with each Mamba block containing the core visual Mamba part and a convolutional feed forward network. For simplicity, we only consider the most critical visual Mamba when calculating FLOPs. This integral part encompasses an SSM and three linear layers. We compute its FLOPs as $128nd + 3nd^2$. Therefore, the FLOPs of the several Inner Mamba blocks in all and an Outer Mamba block in an MiM block can be calculated as $128mnc+3mnc^2$ and 
$128nd+3nd^2$, where $m$ is the number of visual words in a visual sentence and $c$ is the word embedding dimension. Thus, the total FLOPs for the MiM block $FLOPs_{MiM}$ sum to $128mnc+128nd+3mnc^2+3nd^2$, still maintaining a linear complexity. While the FLOPs of the standard transformer block $FLOPs_{ST}$ is $2nd(6d+n)$ \cite{han2021transformer}. Considering that $c \ll d$ and $m \ll n$ in high resolution infrared images, the ratio of $FLOPs_{MiM}$ and $FLOPs_{ST}$ approaches 0, meaning that our MiM block introduces a trivial FLOP increase while offering a superior accuracy-efficiency balance demonstrated in subsequent experiments.

\section{Experiments}
\subsection{Experimental Settings}
\subsubsection{Datasets.}
We choose NUAA-SIRST \cite{dai2021asymmetric} and IRSTD-1k \cite{zhang2022isnet} as benchmarks for training, validation, and testing. NUAA-SIRST contains 427 infrared images of various sizes while IRSTD-1k consists of 1,000 real infrared images of $512\times512$ in size. We resize all images of NUAA-SIRST to size $512\times512$ for training and testing. IRSTD-1k is a more difficult ISTD dataset with richer scenarios and has 1000 infrared images. For each dataset, we use 80$\%$ of images as the training set and 20$\%$ as the test set.

\subsubsection{Evaluation Metrics.}
%We adopt the same evaluation metrics as most prevalent ISTD methods \cite{chen2024tci,zhang2022isnet} and use pixel-level and object-level metrics. The pixel-level metrics include Intersection over Union ($IoU$) and Normalized Intersection over Union ($nIoU$), while the object-level metrics include Probability of Detection ($P_{d}$) and False-Alarm Rate ($F_{a}$). 

We compare our method with other SOTA methods in terms of both pixel-level and object-level evaluation metrics. The pixel-level metrics include Intersection over Union ($IoU$) and Normalized Intersection over Union ($nIoU$), while the object-level metrics include Probability of Detection ($P_{d}$) and False-Alarm Rate ($F_{a}$).

$IoU$ measures the accuracy of detecting the corresponding object in a given dataset. $nIoU$ is the normalization of $IoU$, which can make a better balance between structural similarity and pixel accuracy of infrared small targets. $IoU$ and $nIoU$ are defined as:
\begin{equation}
IoU=\frac{A_{i}}{A_{u}},nIoU=\frac{1}{N}\sum_{i=1}^{N}(\frac{TP[i]}{T[i]+P[i]-TP[i]}),\\
\end{equation}
where $A_{i}$ and $A_{u}$ are the areas of intersection and union region between the prediction and ground truth, respectively. $N$ is the total number of samples, $TP[.]$ is the number of true positive pixels, $T[.]$ and $P[.]$ is the number of ground truth and predicted positive pixels. 

 %Specifically, $nIoU$ is the normalized $IoU$, which can make a better balance between structural similarity and pixel accuracy of infrared small targets. 
$P_{d}$ calculates the proportion between the number of correctly predicted targets $N_{pred}$ and all targets $N_{all}$. $F_{a}$ refers to the ratio of falsely predicted target pixels $N_{false}$ and all the pixels in the infrared image $N_{all}$. $P_{d}$ and $F_{a}$ are calculated as follows:
\begin{equation}
P_{d}=\frac{N_{pred}}{N_{all}},F_{a}=\frac{N_{false}}{N_{all}}.\\
\end{equation}
The correctness of the prediction depends on whether the centroid distance between the predicted target and the ground truth is less than 3 pixels.

\subsubsection{Optimization.}
 The algorithm is implemented in Pytorch, with Adaptive Gradient (AdaGrad) as the optimizer with the initial learning rate set to 0.06 and weight decay coefficient set to 0.0004. 2 NVIDIA A6000 GPU is used for training, with batch size set to 32. Dice loss \cite{sudre2017generalised} is adopted as the loss function. Training on SIRST and IRSTD-1k takes 1000 epochs and 800 epochs respectively.

\begin{table*}
\caption{Accuracy comparison on NUAA-SIRST and IRSTD-1k.  The figures in bold and underlined mark the highest and the 2nd highest ones in each column.} \label{table1}
\centering
\resizebox{1.99\columnwidth}{!}{
    \begin{tabular}{c|c|cccccccc}
    %\toprule
    \toprule[1pt]
        \multirow{2}{*}{Method} & \multirow{2}{*}{Type} &\multicolumn{4}{c}{NUAA-SIRST} & \multicolumn{4}{c}{IRSTD-1k}\\
        & &IoU $\uparrow$ &  nIoU $\uparrow$ &Pd $\uparrow$ &Fa $\downarrow$ & IoU $\uparrow$ &  nIoU $\uparrow$ &Pd $\uparrow$ &Fa $\downarrow$\\
        \midrule
        %\hline
        %Tophat \cite{rivest1996detection}  & Trad& 7.140   & 5.20  & 79.84 & 10.12 & 10.06   & 7.44 & 75.11& 1432\\
        NRAM  \cite{zhang2018infrared} & Trad& 12.16  & 10.22  & 74.52 & 13.85 & 15.25   & 9.899 & 70.68& 16.93\\
        TLLCM \cite{yao2018coarse}  & Trad& 1.029   & 0.905  & 79.09 & 5899 & 3.311   & 0.784 & 77.39& 6738\\
        PSTNN \cite{zhang2019infrared}  & Trad & 22.40   & 22.35 &77.95 & 29.11  & 24.57   & 17.93 & 71.99 & 35.26\\
        MSLSTIPT \cite{sun2020infrared}  & Trad & 10.30   & 9.58  & 82.13 & 1131 & 11.43   & 5.93 & 79.03 & 1524\\

        MDvsFA \cite{wang2019miss}  & CNN & 60.30   & 58.26  &89.35 & 56.35 & 49.50   & 47.41 &82.11 &80.33\\
        ACM \cite{dai2021asymmetric}  & CNN & 72.33   & 71.43 &96.33 & 9.325 & 60.97   & 58.02 & 90.58 & 21.78\\
        AlcNet \cite{dai2021attentional} & CNN & 74.31   & 73.12 & 97.34 & 20.21 & 62.05   & 59.58 & 92.19 & 31.56\\
        DNANet \cite{li2022dense} & CNN & 75.27   & 73.68 & 98.17& 13.62 & 69.01   & 66.22 & 91.92 & 17.57\\
        DCFR-Net \cite{fan2024diffusion} & CNN &  76.23  &74.69 & 99.08& 6.520 &  65.41  &65.45 & 93.60 & \textbf{7.345}\\
        AGPCNet \cite{zhang2023attention} & CNN &  70.60  & 70.16 & 97.25& 37.44 & 62.82   & 63.01& 90.57 &29.82 \\
        Dim2Clear \cite{zhang2023dim2clear}  & CNN& 77.20   & 75.20 &99.10 & 6.72  & 66.3   & 64.2 & 93.7& 20.9\\
        FC3-Net \cite{zhang2022exploring} & CNN& 74.22   & 72.64 &99.12 & 6.569  & 64.98   & 63.59 & 92.93& 15.73\\
        IAANet \cite{wang2022interior}  & CNN-ViT& 75.31   & 74.65 & 98.22&  35.65 & 59.82   & 58.24 & 88.62& 24.79\\
        RKformer \cite{zhang2022rkformer} & CNN-ViT&77.24   & 74.89  & 99.11& \textbf{1.580}  & 64.12  & 64.18 &93.27 & 18.65\\
        ISNet \cite{zhang2022isnet} & CNN& 80.02  & 78.12  & 99.18 &4.924 & 68.77   & 64.84 & 95.56& 15.39\\
        SegFormer \cite{xie2021segformer}& ViT&  67.64  & 66.43 & 89.92& 35.83 & 60.12  & 57.23 & 88.92& 38.93\\
        %Visual Mamba \cite{liu2024vmamba}& Mamba&76.33  &  73.60  & 98.17 & 8.125    &  65.38  & 60.51 & 91.66& 21.08\\
        TCI-Former \cite{Chen_Tan_Chu_Wu_Liu_Yu_2024}& CNN-ViT& \underline{80.79} & \underline{79.85}&\underline{99.23} &4.189 & \underline{70.14} & \underline{67.6}9& \underline{96.31} & 14.81\\
        MiM-ISTD & Mamba & \textbf{80.92} & \textbf{80.13}&\textbf{100} &\underline{2.168} & \textbf{70.36} & \textbf{68.05}& \textbf{96.95} & \underline{13.38}\\
        %\bottomrule
        \bottomrule[1pt]
    \end{tabular}}
    
\end{table*}

\subsection{Quantitative Comparison}

\subsubsection{Accuracy Comparison.} We select some of the SOTA ISTD methods for comparison. As shown in Table~\ref{table1}, our MiM-ISTD generally achieves the best performances in terms of pixel-level metrics and object-level metrics on both datasets. 

%For the pixel-level metrics ($IoU, nIoU$), all the deep-learning methods gain substantial superiority over the traditional methods because deep-learning methods no longer rely heavily on prior knowledge and handcraft features as traditional methods do. Notably, the value of the $IoU$ index is often higher than the $nIoU$ index for all methods. This is because the $IoU$ index pays more attention to small targets, while the $nIoU$ index focuses more on larger targets. Our method achieves the best performance on both $IoU$ and $nIoU$, meaning that our method (1) reaches a balance between the detection of both small and larger targets and (2) achieves the best shape-aware segmentation performance. 

For the pixel-level metrics ($IoU, nIoU$), our method achieves the best performances, thanks to the further integration of Inner Mamba blocks for local feature modelling. In this way, some distinguishable details that may get ignored by the Outer Mamba block can get noticed, which promotes detection accuracy. 

For the object-level metrics ($P_{d}, F_{a}$), how to reach a trade-off between $P_{d}$ and $F_{a}$ is challenging because higher $P_{d}$ also increases the possibility of higher $F_{a}$. From Table~\ref{table1}, we can see that our method generally achieves the best object-level metrics results, especially detecting all the small targets ($P_{d}=100\%$) in the NUAA-SIRST test set. The result demonstrates that our MiM-ISTD can learn better representations to overcome missed detection and false alarms.

\subsubsection{Efficiency Comparison.}
We also compare the efficiency of different methods in terms of parameter number (M), FLOPs (G) and inference time (s) and GPU memory usage (M) during training on $512 \times 512$ infrared image datasets, as shown in Table~\ref{table_complex}. Compared with other methods except ACM, our MiM-ISTD has significantly fewer parameters, GFLOPs, inference time, and memory usage. This is because we adopt the efficient Mamba structure and also use a shared network to calculate the relations of each visual word in visual sentences so that the increased parameters and GFLOPs are negligible. Even compared with the most lightweight ISTD model ACM, our MiM-ISTD still has fewer GFLOPs and significantly higher accuracy. In addition, in higher resolution infrared image scenarios, we show in Fig.~\ref{efficiency} that MiM-ISTD's superiority of inference time and GPU memory usage will further be expanded while still maintaining a superior accuracy.
Notably, compared with the version without Inner Mamba blocks, we notice a slight decrease in model complexity, parameter count, and GPU memory usage. However, the inference speed remains unchanged, and there is a notable decline in average accuracy. This suggests that while the model becomes marginally lighter without the inner Mamba blocks, this comes at the expense of its overall accuracy. Generally, MiM-ISTD reaches the best efficiency-accuracy balance.

\begin{table*}[!t]
    \caption{Comparison of the model parameters (M), FLOPs (G), inference time (s) per image, and GPU memory (M) per 4 batch size of different methods.} 
\centering
\resizebox{1.99\columnwidth}{!}{
    \begin{tabular}{c||cccc||c}
    \toprule[1pt]
        Method & Param(M) $\downarrow$& FLOPs(G) $\downarrow$ & Inference(s) $\downarrow$ & Memory(M) $\downarrow$& Avg nIoU $\uparrow$\\
        %\cline{2-9}
        \midrule
        %\hline
        %MDvsFA \cite{wang2019miss}  & 3.77& 834&0.24 & 10248M-10bs &\\
        ACM \cite{dai2021asymmetric}  & 0.52& 2.02&0.01 & 1121 & 64.73\\ % memory test 785
        DNANet \cite{li2022dense} & 4.7 & 56.34 &0.15  & 10617&69.95\\
        %AGPCNet   & 48.48& &0.08 &\\
        IAANet \cite{wang2022interior}  & 14.05 & 18.13&0.29 & 45724&66.45\\
        RKformer \cite{zhang2022rkformer} & 29.00 & 24.73&0.08 &-&69.54\\
        ISNet \cite{zhang2022isnet} & 1.09& 122.55&0.05 &15042&71.48\\
        TCI-Former \cite{Chen_Tan_Chu_Wu_Liu_Yu_2024}& 3.66 & 5.87 & 0.04 & 5160&73.77\\
        w/o Inner Mamba & 4.67 & 3.91 & 0.03 & 1434&70.86\\
        %MiM-ISTD & 1.16 & 1.01 & 0.03 & 1774&74.01\\ % memory test 956
        MiM-ISTD & 4.76 & 3.95 & 0.03 & 1996&74.09\\
        \bottomrule[1pt]
    \end{tabular}}
\label{table_complex}
\end{table*}

\begin{figure*}
    \centering    \includegraphics[width=0.96\textwidth,height=0.76\textwidth]{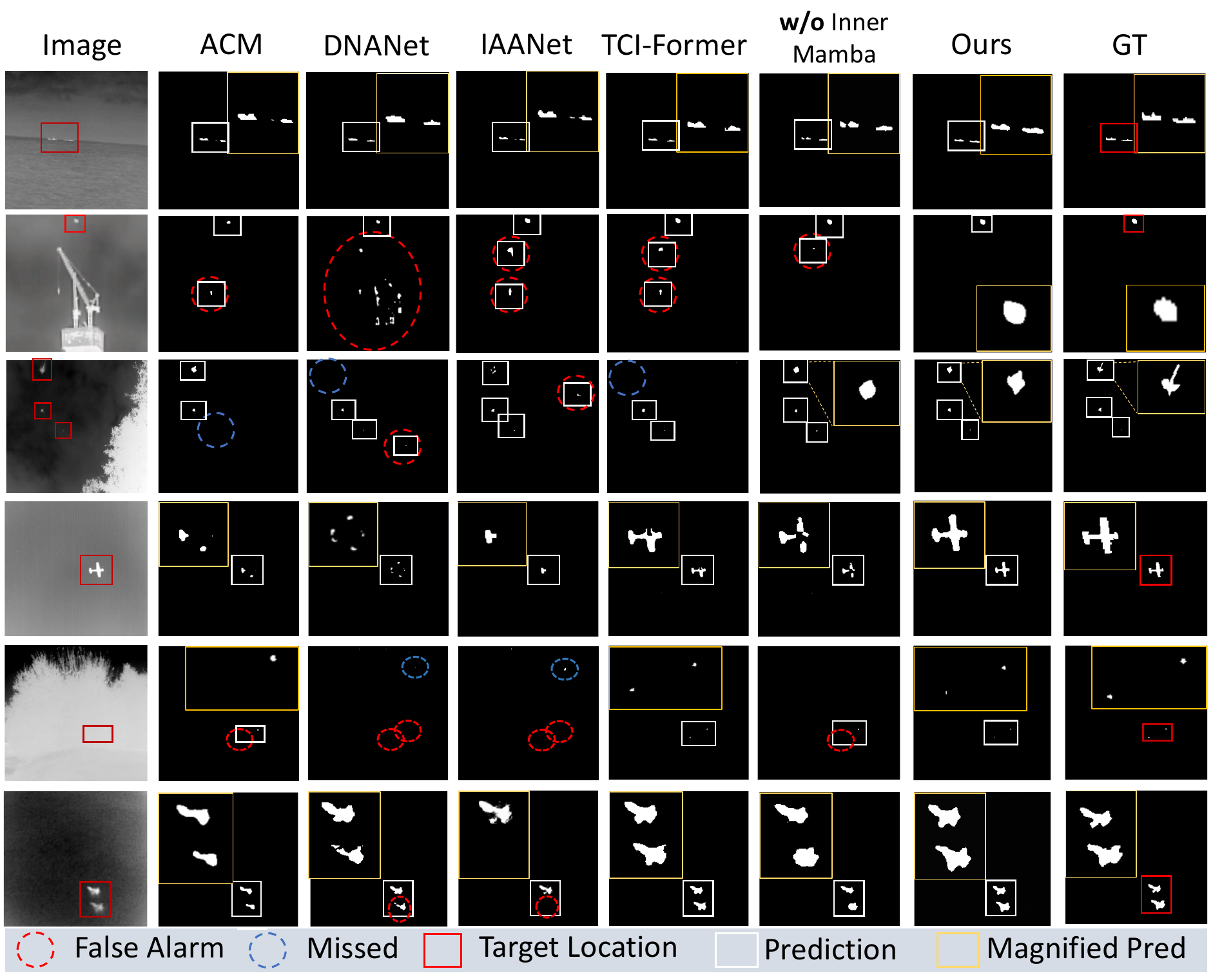}
    \caption{Visualization of predicted mask results. The predicted targets are amplified in the yellow boxes for clearer observation, and we highlight some inaccuracies, such as false alarms and missed detection, made by other methods.}\label{vis}
\end{figure*}

\subsection{Visualization}

\subsubsection{Visualization of Mask Results.} Visual results with closed-up views of different methods are shown in Fig.~\ref{vis}, where present methods more or less suffer from incomplete detection and missed detection. Compared with other SOTAs, our MiM-ISTD better curtails these cases and more completely detects the shapes of all small targets. This is because integrating an Inner Mamba block assists our network to further exploit more local features, which promotes more refined detection of small targets. This can also be proved by comparing MiM-ISTD visual results with the "no Inner Mamba" visual results, where abandoning the Inner Mamba brings worse detection performances.

\subsubsection{Visualization of Feature Maps.} 
\begin{figure*}
\centering
\subfloat[]{\includegraphics[width=0.70\textwidth,height=0.34\textwidth]{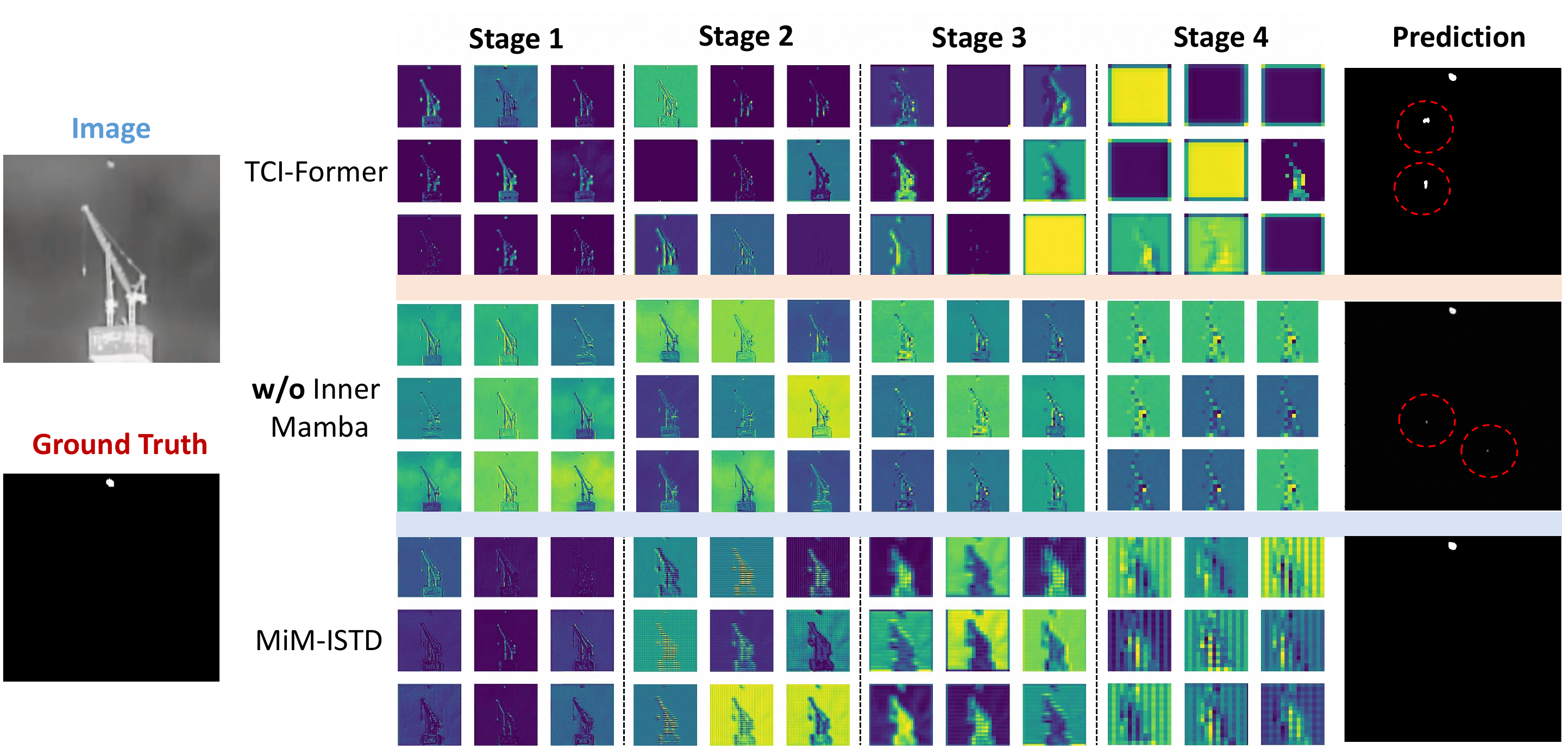}%
\label{feat_vis}}
\hfil
\subfloat[]{\includegraphics[width=0.29\textwidth,height=0.30\textwidth]{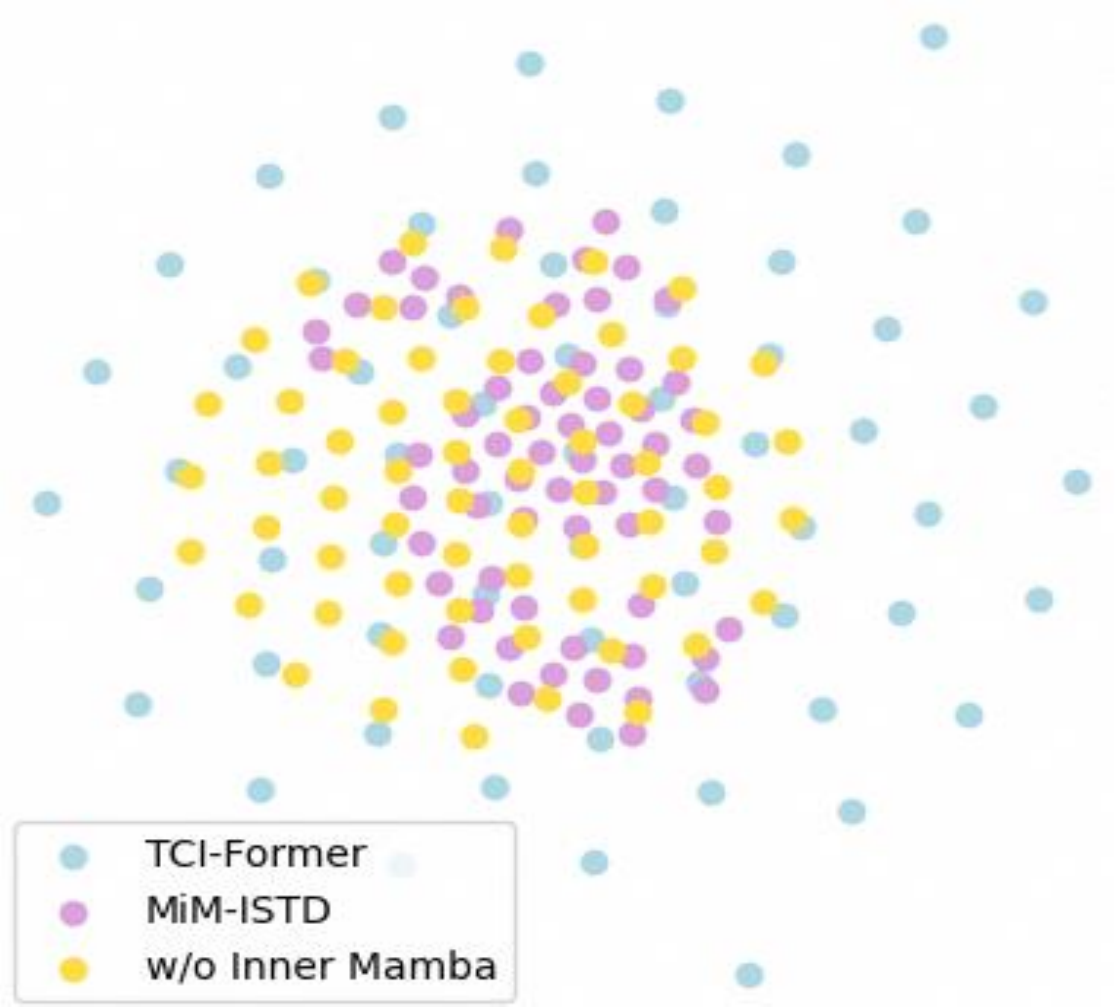}%
\label{tsne}}
\caption{(a) Feature map outputs of Stage 1/2/3/4.
 and (b) Visualization through T-SNE of feature outputs from the third stage, across 64 channels, using different methods.}
\end{figure*}

We visualize the learned features of TCI-Former and MiM-ISTD to further
understand the effect of the proposed method. The feature maps are formed by reshaping the patch embeddings according to their
spatial positions. The feature map outputs of stages 1, 2, 3, and 4 are shown in Fig.~\ref{feat_vis}, where 9 feature map channels are randomly sampled for each of these outputs. In MiM-ISTD, the local information is better preserved in deeper layers compared to TCI-Former. Also, MiM-ISTD has higher feature consistency among each channel than TCI-Former \cite{Chen_Tan_Chu_Wu_Liu_Yu_2024}, the present SOTA method, meaning that the features extracted by MiM-ISTD are more focused on the target. These benefits are owed to the introduction of the Inner Mamba block to further model local features. We additionally visualize 64 channels of the stage 3 feature output using t-SNE \cite{van2008visualizing} in Fig.~\ref{tsne} to demonstrate our analysis. We observe that the features of MiM-ISTD are more concentrated than the model without Inner Mamba blocks and the present SOTA method \cite{Chen_Tan_Chu_Wu_Liu_Yu_2024}. This observation aligns with the results obtained from feature comparison. In general, our MiM-ISTD exhibits stronger discriminative power. %This benefit owe to the introduction of inner Mamba block to further model local features.

% \begin{figure*}
%     \centering    \includegraphics[width=0.99\textwidth,height=0.34\textwidth]{mim_eccv/feat_vis.pdf}
%     \caption{Visualization of the features of TCI-Former and MiM-ISTD.}\label{feat_vis}
% \end{figure*}

\begin{figure*}
    \centering    \includegraphics[width=0.95\textwidth,height=0.38\textwidth]{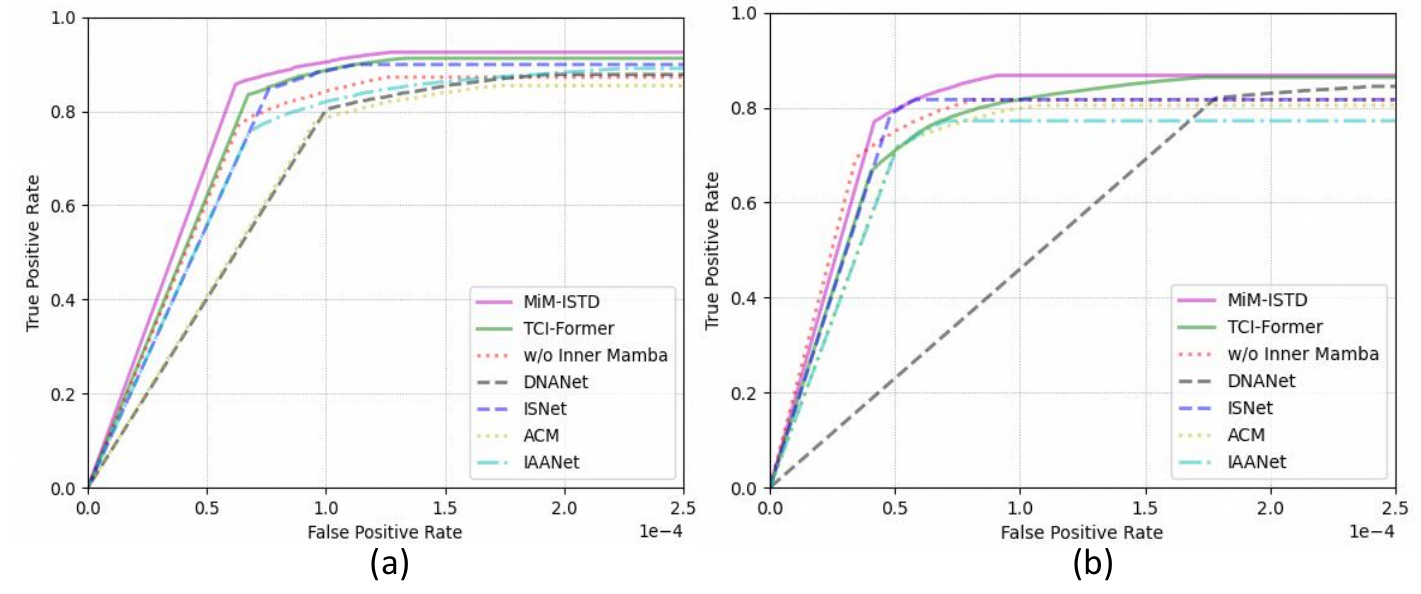}
    \caption{ROC curves on (a) NUAA-SIRST and (b) IRSTD-1k datasets.}\label{roc}
\end{figure*}

\subsection{ROC results}
While $IoU$, $nIoU$, $P_{d}$ and $F_{a}$ measure the segmentation performance under a fixed threshold, the ROC can provide an overall evaluation under multiple different thresholds. We also compare the ROCs among other SOTA methods in Fig.~\ref{roc}. It can be noted that our MiM-ISTD exhibits overall impressive performance on both datasets, particularly within intervals of low false positive rates where its true positive rate swiftly escalates, demonstrating robust detection capabilities. Additionally, at higher false positive rates, MiM-ISTD maintains a relatively high true positive rate, indicating good overall robustness. The results further demonstrate the superiority of our method over other SOTA methods.

\subsection{Ablation study}
\subsubsection{Ablation study of Each Module.} 
The ablation study of each module is shown in Table~\ref{ablation1}. Our baseline (No.$\#1$) uses the plain visual Mamba \cite{liu2024vmamba} encoder as our encoder with descending spatial resolutions, where no Inner Mamba block is employed. We also replace the Inner Mamba block of our MiM-ISTD with the standard convolution, batch normalization and activation operations (No.$\#2$) to examine the effect of our Inner Mamba block. We find that our present setting brings the best results, because convolution typically focus on local information, while the Inner Mamba block can capture complex spatial relationships and integrate multi-directional information to have a more delicate perception and identification of small targets.

\begin{table}
\centering
\caption{Ablation study of each module on NUAA-SIRST.}
\resizebox{0.99\columnwidth}{!}{
    \begin{tabular}{c|c|ccccc}
    \toprule
        \multirow{2}{*}{No.}&\multirow{2}{*}{Method} & \multicolumn{4}{c}{NUAA-SIRST}\\
        %\cline{2-9}
        & &IoU $\uparrow$ & nIoU $\uparrow$ &Pd $\uparrow$ &Fa $\downarrow$\\
        \midrule
        %\hline
    $\#$1& Baseline &76.33  &  73.60  & 98.17 & 8.125 \\
    $\#$2& Inner Mamba$\rightarrow$DWConv  &  78.47  & 76.26 & 99.15 & 5.014\\
    $\#$3& Ours & \textbf{80.92} & \textbf{80.13}&\textbf{100} &\textbf{2.168}\\
        \bottomrule
    \end{tabular}}
 \label{ablation1}
\end{table}

\begin{table}
\centering
\caption{Ablation study of the granularity of patch division on NUAA-SIRST.} \label{ablation2}
\resizebox{0.99\columnwidth}{!}{
    \begin{tabular}{c|c|ccccc}
    \toprule
        \multirow{2}{*}{No.}&\multirow{2}{*}{Method} & \multicolumn{4}{c}{NUAA-SIRST}\\
        %\cline{2-9}
        & &IoU $\uparrow$ & nIoU $\uparrow$ &Pd $\uparrow$ &Fa $\downarrow$\\
        \midrule
        %\hline
    $\#$1&VW $\rightarrow 1 \times 1$  & 78.71   & 76.43 & 99.12& 5.965\\
    $\#$2&VW $\rightarrow 4 \times 4$  &  79.50  & 77.68 & 100 & 2.980\\
    \midrule
    $\#$3&VS $\rightarrow 4 \times 4$ &  79.67& 78.04 & 100 &  3.685\\
    $\#$4&VS $\rightarrow 16 \times 16$ & 69.32 & 67.58 & 93.52 & 22.21 \\
    \midrule
    $\#$5& Ours (VW $\rightarrow 2 \times 2$, VS $\rightarrow 8 \times 8$ ) & \textbf{80.92} & \textbf{80.13}&\textbf{100} &\textbf{2.168}\\
        \bottomrule
    \end{tabular}}
\end{table}

\subsubsection{Ablation Study of the Granularity of Patch Division.} To explore the impact of information contained in each visual word produced from the convolutional stem from the start, we adjust the representation granularity of sub-patches so that each visual word corresponds to a $1\times1$ (No.$\#1$) or $4\times4$ (No.$\#2$) pixel region in the original image, in contrast to $2\times2$ of our present setting. We also fix the $2\times2$ reception field of visual word and change the initial spatial shape of visual sentence from $\frac{H}{8} \times \frac{W}{8}$ to $\frac{H}{4} \times \frac{W}{4}$ (No.$\#3$) and $\frac{H}{16} \times \frac{W}{16}$ (No.$\#4$) to ablate the effect of visual sentence granularity. We can observe that too large or too small division granularity cannot bring the most ideal performance. We adopt the configuration that demonstrates the best performance in our present setting.

\section{Conclusion}
In this paper, we propose a Mamba-in-Mamba (MiM-ISTD) structure for efficient ISTD. We uniformly divide the image into patches as visual sentences, and further split each patch to multiple smaller sub-patches as visual words. We devise a pure Mamba-based MiM hierarchical encoder that encompasses stacked MiM blocks. Each MiM block contains an Outer Mamba block to process the sentence embeddings and an Inner Mamba block to model the relation among word embeddings. The visual word embeddings are added to the visual sentence embedding after a linear projection. Experiments show that our method can achieve more efficient modelling of both local and global information. %Experiments on two datasets have demonstrate the effectiveness of our method.

%\section*{Acknowledgments}
%This should be a simple paragraph before the References to thank those individuals and institutions who have supported your work on this article.


\bibliographystyle{IEEEtran}
\bibliography{egbib}

\newpage
\begin{comment}
\section{Biography Section}
If you have an EPS/PDF photo (graphicx package needed), extra braces are
 needed around the contents of the optional argument to biography to prevent
 the LaTeX parser from getting confused when it sees the complicated
 $\backslash${\tt{includegraphics}} command within an optional argument. (You can create
 your own custom macro containing the $\backslash${\tt{includegraphics}} command to make things
 simpler here.)
 
\vspace{11pt}

\bf{If you include a photo:}\vspace{-33pt}
\begin{IEEEbiography}[{\includegraphics[width=1in,height=1.25in,clip,keepaspectratio]{fig1}}]{Michael Shell}
Use $\backslash${\tt{begin\{IEEEbiography\}}} and then for the 1st argument use $\backslash${\tt{includegraphics}} to declare and link the author photo.
Use the author name as the 3rd argument followed by the biography text.
\end{IEEEbiography}

\vspace{11pt}

\bf{If you will not include a photo:}\vspace{-33pt}
\begin{IEEEbiographynophoto}{John Doe}
Use $\backslash${\tt{begin\{IEEEbiographynophoto\}}} and the author name as the argument followed by the biography text.
\end{IEEEbiographynophoto}
\end{comment}

\vfill

\end{document}